\documentclass{article}
\usepackage[square,numbers]{natbib}
\usepackage{booktabs}
\usepackage{multirow,array}
\usepackage{amsmath}
\usepackage{graphicx}

\usepackage[final]{mlncp_2023}

\usepackage[utf8]{inputenc} 
\usepackage[T1]{fontenc}    
\usepackage{hyperref}       
\usepackage{url}            
\usepackage{booktabs}       
\usepackage{amsfonts}       
\usepackage{nicefrac}       
\usepackage{microtype}      
\usepackage{xcolor}         

\title{Neuromorphic Co-Design as a Game}

\author{%
  Craig M.~Vineyard, William M. Severa, $\&$ James B. Aimone \\
  Center for Computing Research\\
  Sandia National Laboratories\\
  Albuquerque, NM 87185 \\
  \texttt{\{cmviney, wmsever, jbaimon\}@sandia.gov} \\
}

\begin{document}

\maketitle

\begin{abstract}
Co-design is a prominent topic presently in computing, speaking to the mutual benefit of coordinating design choices of several layers in the technology stack. For example, this may be designing algorithms which can most efficiently take advantage of the acceleration properties of a given architecture, while simultaneously designing the hardware to support the structural needs of a class of computation. The implications of these design decisions are influential enough to be deemed a lottery, enabling an idea to win out over others irrespective of the individual merits. Coordination is a well studied topic in the mathematics of game theory, where in many cases without a coordination mechanism the outcome is sub-optimal. Here we consider what insights game theoretic analysis can offer for computer architecture co-design. In particular, we consider the interplay between algorithm and architecture advances in the field of neuromorphic computing. Analyzing developments of spiking neural network algorithms and neuromorphic hardware as a co-design game we use the Stag Hunt model to illustrate challenges for spiking algorithms or architectures to advance the field independently and advocate for a strategic pursuit to advance neuromorphic computing.
\end{abstract}

\section{Introduction}
The field of computing has seen great advances from algorithms and architectures down to materials and devices over decades of innovation and optimization ~\cite{hennessy2019new,hennessy2017computer}. With an eye towards even more sophisticated performance, co-design is readily being considered across the technology stack. The sentiment being, that further advances can be achieved by considering multiple, interrelated design facets simultaneously. In this manner, by designing features and functionality jointly, the combined outcome will be greater than pursuing individual optimizations. For example, this may be algorithms which can most efficiently take advantage of the acceleration properties of a given architecture or vise versa tailoring architectural optimizations to more efficiently execute facets of classes of algorithms of interest. 

Even in the absence of explicit co-design, historical advancements have not been in isolation but intrinsically have been iterative design progressions. For example, the identification of important instructions to enable (whether in explicit hardware support, instruction representation, or other means) is defined by the algorithms the instructions represent. Analogously, letters of the alphabet have unique usage distributions in relationship to vocabulary. The field of information theory quantifies and exploits this principle for efficient encoding, but requires a language model to indicate how letters are frequently composed as words (the distributions) to then create an efficient encoding. Likewise, the operations which compose algorithms of interest can guide the optimization of computational designs both in terms of the definition of the instruction set as well as in designing hardware to efficiently execute important computations.   

\begin{figure}
  \centering
  \includegraphics[height=1.5in]{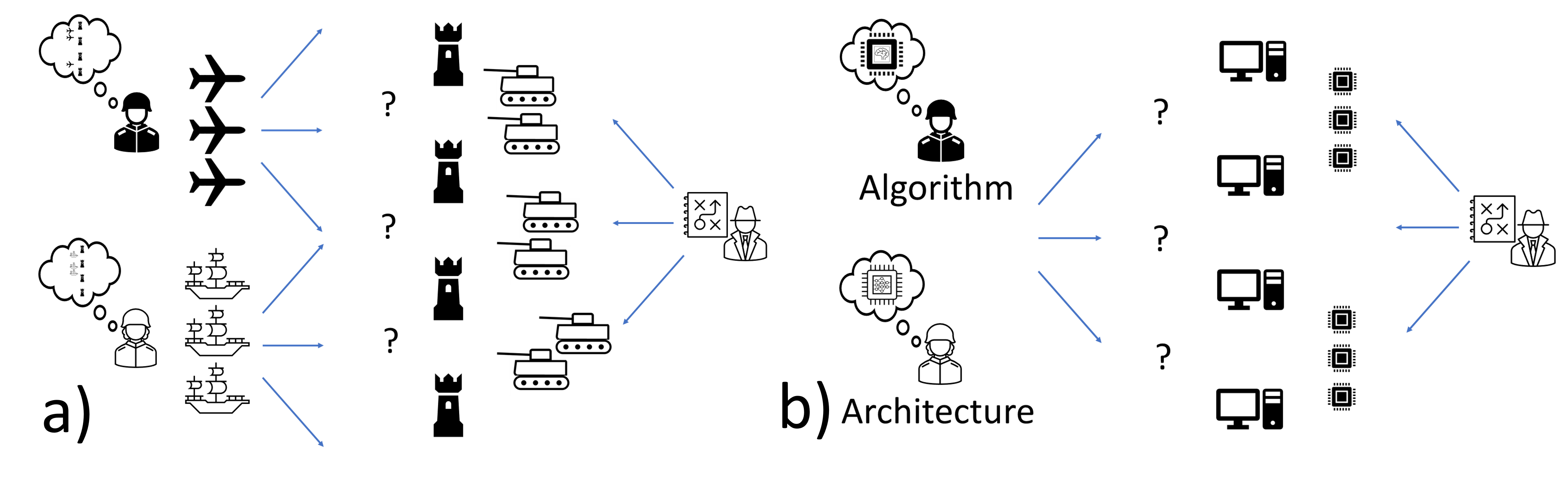}
  \label{fig:coordinateGame}
  \vspace{-6mm}
  \caption{Conceptual portrayal of a joint decision scenario where two decision makers combine forces to take upon an opponent. In a) the outcome depends upon the force allocations of two military leaders at the battle sites in the middle. A coordinated effort entails the decision makers work together to maximize their effort. Analogously, b) extends this scenario to consider the strategic co-design in computing. In this manner rather than military leaders allocating forces, how algorithms and architectures are designed influence their performance on computational workloads.}
  \vspace{-6mm}
\end{figure}

The entanglement of computational design choices spans the history of computing and in fact, as identified by the `Hardware Lottery' impacts the perceived superiority of ideas over alternatives ~\cite{hooker2021hardware}. Namely, the Hardware Lottery study showcases how one algorithm may be deemed superior to another due to enabling hardware rather than the superiority of the algorithm itself (with a software lottery also impacting idea comparisons). And notably, an illustrative exemplar in Hooker's historical narrative is how neural networks rose to prominence with the enablement of Graphic Processing Unit (GPU) acceleration even though algorithmic underpinnings like backpropagation were around earlier. 
We consider computational co-design to be more than a lottery -- a game. 
The field of game theory is the mathematical analysis of strategy. Accordingly, it offers many formulations for analyzing player interactions where decisions are interdependent. In an optimization problem, the goal is to determine the parameter values which maximize or minimize an objective. In game theory this optimization is dependent upon the decisions of more than one player as depicted by Fig. ~\ref{fig:coordinateGame}. Accordingly, we see it as well suited for considering the joint decision making of the co-design of computational algorithms and architectures. 

In particular, we consider the field of neuromorphic computing. Novel algorithm formulations are actively being pursued across a range of applications seeking to find more efficient ways to perform computations using neurons as the core computational element~\cite{schuman2022opportunities,aimone2022review,aimone2019neural}. And likewise, novel computational architectures are considering how to best structure processing elements, communication, and memory while looking to biological brains for inspiration~\cite{furber2016large,bouvier2019spiking,schuman2017survey,christensen20222022}. We note other facets of the technology stack are also readily exploring neuromorphic computing advances such as how to utilize novel materials to develop efficient devices for composing neuromorphic architectures. For our work here, we are focusing upon the neuromorphic co-design of algorithms and architectures. In doing so, we strive to understand implications for the neuromorphic field. Do we need known spiking neural algorithms whose theoretical promise can justify architectural instantiation? Or can novel architectures precede algorithmic theory and spur innovation? Can the latter be pursued without skewing the path forwards given the known Hardware Lottery effect? As follows, we seek to explore the implications of neuromorphic co-design through a game theoretic modeling and analysis. First we provide a brief background of how the field of game theory represents some related scenarios, then we present neuromorphic co-design via the lens of the Stag Hunt game and analyze scenarios to offer insight into the implications for the neuromorphic field.

\section{Background}

The mathematics of game theory are applicable for examining a wide range of strategic interactions ~\cite{von1947theory}. This includes competitive as well as cooperative interactions, and can range from individual to population level models. Additionally, there are multiple solution concepts which equate to providing a strategy, identifying the best actions to take for the problem ~\cite{osborne1994course,karlin2017game,dutta1999strategies,roughgarden2010algorithmic}.  

Accordingly, there are many ways in which game theory can offer insight to an interaction depending upon the problem formulation. 
For example, with respect to technology development, in a competitive scenario there are Research and Development (R$\&$D) models in which one corporation may want to beat out a competitor and secure the market share~\cite{dutta1999strategies}. In this manner the problem formulation has a temporal component of making a research investment in the aspiration to secure a profit in the future and justifying the research cost. An example application might be a pharmaceutical investment where winning a patent gives the prevailing corporation an advantage and their advantage comes at the expense of the opposing corporations. 

Population level and evolutionary game models can examine strategic interactions in a broader context. This level analysis can bring insight into how individual decisions can impact the whole. For example, Braees's paradox illustrates that how infrastructure improvements can lead to decreased system functionality ~\cite{karlin2017game}. In this sense, the impact of technology development may require a broader view to comprehend the true impact. And evolutionary game theory explores concepts such as whether a new strategy can be introduced into an existing population or will be pushed out. Perspectives such as this may readily offer insight into what conditions are necessary for new technology success. 

Cooperative games model interactions where instead of players succeeding at the expense of the opponent, outcomes are the consequence of coordinated actions. As touched upon there are several manners in which game theoretic analysis can offer insight into facets of the development of neuromorphic computing. For the purpose of examining co-design, we will consider a cooperative game theory model here as described next. 

\section{Modeling and Analysis}

Here we consider equilibrium analysis to offer insight into the implications for neuromorphic computing co-design. Equilibrium analysis indicates what strategies players cannot improve upon without changes to the game structure or an opponent's strategy. 

For our modeling and analysis we examine neuromorphic computing co-design as a Stag Hunt game. The dynamics captured by a Stag Hunt game represent the scenario where two hunters can team up to bring in the higher-value stag, or independently can secure a less rewarding hare ~\cite{dutta1999strategies,easley2010networks}. Attaining the higher reward is dependent upon the coordination of the two hunters/players. The general representation of a Stag Hunt game is captured by the payoff matrix shown in Table \ref{table:StagHunt}. In this context, two players $X$ and $Y$ are represented by the rows and columns respectively. Each sub row or column corresponds to the actions the respective player takes. And the intersection of actions within the table provides the utility value each player receives as a result of that joint outcome. In this case, the players select their actions simultaneously, and a strategy is a policy which dictates the selection of an action to take (for example a distribution over the actions). 

The core structure of the payoff values in a Stag Hunt game is such that $a$ > $b$ $\geq$ $d$ > $c$. Subtle scenarios within this general structure exist such as whether the reward for pursuing a Hare is shared when both players select that action as compared with receiving the full reward under the miss-aligned action selection scenarios. Preserving the nominal relationship between these reward values impacts the dynamics of the interaction and differentiates this from other games~\cite{yamamoto2019single}. And in fact, other game structures may also model facets of co-design. However, here we focus upon the insights the Stag Hunt game provides for the co-design challenge facing neuromorphic computing as the field seeks to advance spiking neural algorithms and neuromorphic hardware. To this effect, Table \ref{table:NMCgame} illustrates exemplar utility values for considering Neuromorphic Co-Design as a Stag Hunt game which we will examine shortly. Notably, the Stag reward (SNN-SNN action pair) has the highest payoff for both players but depends upon the action selected by the other player. Alternatively, Hare (ANN) has a lower payoff but can be attained independently. This problem representation is not stating that ANN efforts are not high value, but is intentionally modeling the potential transformative reward that advancing neuromorphic computing can have. These game dynamics create payoff dominant and risk dominant pure strategy Nash equilibria for each respective action pairing. Additionally, the Stag Hunt game also has a mixed strategy equilibria, where rather than playing either action exclusively, the players play each part of the time. The mixed strategy equilibria solution depends upon the payoff utility ratios and will be the focus of our ensuing analysis momentarily. 

The Stag Hunt game itself has been well studied and our analysis is not novel with respect to the game dynamics. However, here, we use the insights regarding these solutions to inform the implications of neuromorphic computing co-design decisions. In this context, rather than pursuing a Stag or a Hare, the action choices become whether hardware and software designers pursue spiking neural network (SNN) architectures and algorithms respectively. For this problem formulation, we contrast the pursuit of SNNs with conventional Artificial Neural Networks (ANNs). In this manner, ANNs are intended to broadly represent the general taxonomy of deep learning inspired neural networks with less biological fidelity such as convolutional neural networks (CNNs), deep neural networks (DNNs), etc. Our intention here is not to prescribe how much neural-inspired functionality is needed either in terms of algorithms or architectures but to examine the implications as that design consideration underlies the development of novel algorithms and architectures. In this problem formulation, a player is paired with the complementary player type to model co-design decision making - depicted by Table \ref{table:NMCgame}. Note, additional model complexity can allow for the payoff values between the player types to be asymmetric. Alternative game formulations may represent like player types such as two spiking algorithm players whose strategic interaction outcome models some facet of their efficiency in computation. Rather than representing neuromorphic co-design, that sort of game formulation might be applied to explore the selection of a solution for a particular target architecture. 

\begin{table}[]
    \caption{Stag Hunt Payoff Matrix Structure}
    \label{table:StagHunt}
    \parbox{\linewidth}{
        \centering
        \setlength{\extrarowheight}{2pt}
        \begin{tabular}{cc|c|c|}
          & \multicolumn{1}{c}{} & \multicolumn{2}{c}{Player $Y$}\\
          & \multicolumn{1}{c}{} & \multicolumn{1}{c}{$Stag$}  & \multicolumn{1}{c}{$Hare$} \\\cline{3-4}
        \multirow{2}*{Player $X$}  & $Stag$ & $(a,a)$ & $(c,b)$ \\\cline{3-4}
        & $Hare$ & $(b,c)$ & $(d,d)$ \\\cline{3-4}
    \end{tabular}
    }
\end{table}

\begin{table}[]
   \vspace{-3mm}
    \caption{Neuromorphic Co-Design Game}
    \label{table:NMCgame}
    \parbox{\linewidth}{
        \centering
        \setlength{\extrarowheight}{2pt}
        \begin{tabular}{cc|c|c|}
          & \multicolumn{1}{c}{} & \multicolumn{2}{c}{Algorithm Player ($P_{2}$)}\\
          & \multicolumn{1}{c}{} & \multicolumn{1}{c}{$SNN$}  & \multicolumn{1}{c}{$ANN$} \\\cline{3-4}
        \multirow{2}*{Architecture Player ($P_{1}$)}  & $SNN$ & $(5,5)$ & $(1,3)$ \\\cline{3-4}
        & $ANN$ & $(3,1)$ & $(2,2)$ \\\cline{3-4}
    \end{tabular}
    }
    \vspace{-7mm}
\end{table}

Next we explore a few implications from a co-design perspective for neuromorphic computing. The exact values themselves are not intended to capture nuances such as R$\&$D costs or market value. But rather, by considering their relative relationships we can analyze the implications for the influence architectural advances have on algorithm development and vice versa. 

\vspace{-4mm}

\subsection{Neuromorphic Co-Design Analysis}

\subsubsection{Co-Design Mixed Strategy Dilemma} 

Beginning with the utilities represented in Table \ref{table:NMCgame}, this captures the context where architecture and algorithm players can both prosper if they each pursue SNN advances together. Conversely, they can also do reasonably well under the ANN-ANN action pair. This is not to say the joint pursuit of ANN technologies is not advantageous as many exciting research advances have illustrated that very scenario. But rather, we are considering the implications of novel alternative SNN research and innovation. In this regard, the ANN action scenario represents if either player is to pursue the more known outcome of ANN pursuits. For example if the algorithm player were to pursue a breakthrough in DNNs or conversely if the architecture player were to design hardware to more efficiently execute DNNs. Under this co-design game scenario, however, that outcome has a lower utility value as it deviates from the pursuit of innovative SNN research (and to conform to the structure of the Stag Hunt game). 

In addition to the pure strategy solutions where each player is fully committed to their research pursuit (SNNs or ANNs), a mixed-strategy equilibria solution also exists. Various analytic techniques can provide the solutions to games, and their in-depth description are beyond the scope of this paper. For the Stag Hunt game, a simple equation identifying the mixed equilibrium distribution is as follows. If $Player_{2}$ plays SNN with probability $x$ and ANN with probability $1-x$, then $Player_{1}$'s best strategy is when they are indifferent to changing their action distribution. This occurs under the following equation (and for symmetric utility values is equivalent for $Player_{2}$):    \\
\vspace{-\baselineskip}
\begin{center}
\begin{math}
\begin{aligned}
SNN(x) &= ANN(x)  \\
ax + (1-x)c &= bx + (1-x)d \\
ax + c - cx &= bx + d - dx \\
ax -bx -cx +dx &= d - c \\
x(a - b - c + d) &= d - c \\
x &= \frac{d - c}{a - b - c + d}
\end{aligned}
\end{math}
\end{center}

For the base scenario (Table \ref{table:NMCgame}), this equates to a mixed strategy equilibria where both player's action distributions are 0.333 $SNN$ and 0.667 $ANN$. While the promise of SNN may be large, this illustrates the perhaps unexpected scenario that the majority action is to pursue the more well known and less risky ANN action. Given the technical maturity of SNN and ANN technologies one could argue that their respective modeling as Stags or Hares should be switched, in which case this analysis would correspondingly  imply the alternative solution and SNNs would be the majority research pursuit. However, for the co-design scenario here, this modeling and analysis formulation is exploring the implications for co-design dynamics to advance SNNs. And importantly, this illustrates why even if the reward is promising that independent endeavours may not be enough to advance neuromorphic computing. 

\subsubsection{Increasing $SNN$ Value}

Here we analyze the scenario where the reward payoff for $SNN$ increases in order to consider the implication if either the development of an efficient (low power, large scale, fast) neuromorphic architecture emerges and/or the development of an algorithmic breakthrough to produce spiking algorithms in the manner backpropagation has done for ANNs. Can the promise of either breakthrough instigate research in the other domain? In this regard, we imagine the $a$ utility value increasing substantially. 

Counterintuitively, rather than driving the players to pursue the large reward of the $SNN$ action, this actually reduces the mixed strategy distribution so that $SNN$ is rarely played. For example, as shown in Fig. 2 (left), if a 10x increase in reward were considered and all other utilities remain the same (i.e. $a$ = 50), the resulting action distribution becomes 0.0208 $SNN$ and 0.979 $ANN$. Likewise, for a 100x increase in reward and all other utilities remain the same ($a$ = 500), the resulting action distribution becomes 0.002 $SNN$ and 0.998 $ANN$. Effectively, rather than driving innovation in $SNN$ research the game dynamics converge toward entirely pursuing $ANN$. And so simply producing a breakthrough neuromorphic architecture is not enough to usher forwards the pursuit of spiking algorithms or for an algorithm innovation to drive novel neuromorphic architectures. This counterintuitive outcome is a challenge of collaboration. Even though the $SNN-SNN$ outcome is substantially larger it introduces risk as the outcome depends upon the joint action of the opposing player. Whereas instead the $ANN$ pursuit can be achieved individually, independent of the other player. In the context here, the cooperative challenge becomes innovative SNN algorithms needing a neuromorphic architecture to execute their advantage or reciprocally a neuromorphic architecture needing a SNN algorithm to make sure of the innovative hardware. 

\subsubsection{Compromise $SNN$ $\&$ $ANN$}
To model a compromise where the $ANN$ value approaches that of $SNN$, we explore increasing the $b$ utility towards $a$. In this regard, rather than seeking to advance $SNN$ research the compromise is to support the known $ANN$s while still wanting to pursue $SNN$s. 
For example, an architecture may factor design choices such as matrix multiplication to align with supporting $ANN$ calculations as well as appeal towards $SNN$ principles like sparsity or event driven operation. To examine these dynamics, we first raise the $a$ utility value so that $b$ has room to increase while maintaining the Stag Hunt structure defined earlier. Consider the following set of utilities: $a=10$, $c=1$, $d=2$, and $b=3:9$. As $b$ increases, so does the mixed strategy selection of action $SNN$. In this exemplar range, as shown by Fig. 2 (right), the respective $SNN$ strategies are: 0.125, 0.143, 0.167, 0.200, 0.250, 0.333, and 0.500. Note the above outcomes also represent increasing d, and c in alignment with b. Intuitively, for our scenario this makes sense as $d$ corresponds to both players pursuing $ANN$ advances and $b$ is one player doing so unilaterally. 

Due to the structure of the game (the relationship of the utility rewards and their relative values), when additionally a risk dominance ratio is met, it prevents the SNN strategy from exceeding 0.5 ~\cite{harsanyi1988general} effectively capping a drive to pursue $SNN$ over the allure of the known $ANN$ outcome. Intuitively we can see this as $SNN$s and $ANN$s converging in performance. Formally. a risk dominant ratio is the product of the deviation losses and calculates the impact of miss-aligned actions for a given strategy. In other words, how much risk of lost reward is there if the opposing player does not play the expected action. Effectively, this dynamic of the game structure bounds how much risk the player's strategy employs. 

\begin{figure}
  \centering
  \includegraphics[width=\textwidth]{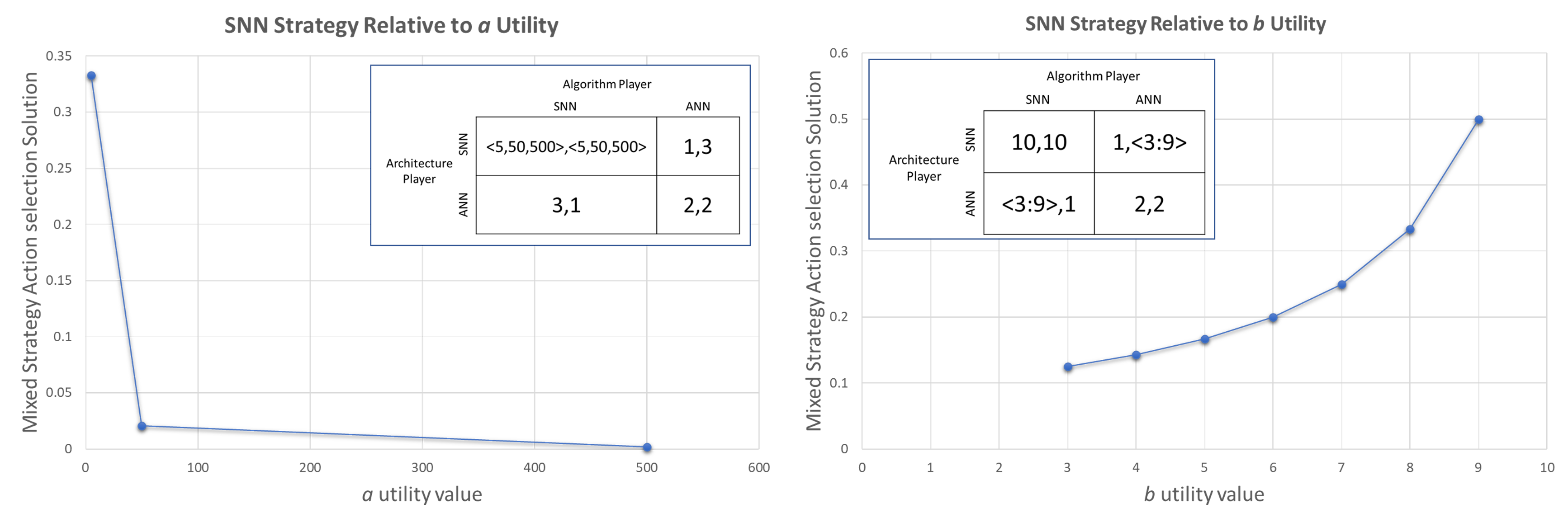}
  \label{fig:simPlots}
  \vspace{-6mm}
  \caption{Mixed strategy solution trends for SNN action selection for the `Increasing SNN Value' scenario (left) and `Compromise SNN $\&$ ANN' scenario (right)}
  \vspace{-6mm}
\end{figure}

\section{Conclusions}
A common computer science analysis technique is to map one problem to another for which properties such as complexity are known. By doing so, the computational difficult of the algorithm being evaluated can be established. Likewise, in game theory there is value in formulating a problem as it relates to a known game structure. The value is not in discovering unknown solutions, but rather this offers insights based upon the analysis of the dynamics of well known games for the problem under consideration. We have done so here, considering the insights the Stag Hunt game offers for considering the challenges of cooperation in the co-design of neuromorphic computing. This first step in bringing awareness to the concept of neuromorphic computing development as a co-design game offers insight into why some of the best intentioned individual technical advances may not have the impact they desire. This perspective also offers understanding into why some of the results we can reflect back upon may have occurred. For example, analysis has shown that simply mapping ANNs/DNNs to spiking neuromorphic architectures may not be the most advantageous algorithm to architecture mapping~\cite{davies2021advancing,daviesNICE2022}, however there is certainly an allure to do so (as our co-design analysis shows). Of course, that is not to say neural networks should not be pursued for neuromorphic hardware. However, the ANN algorithms which have emerged due to other hardware like GPUs, do not exploit the advantages of neuromoprhic hardware. And rather, some benchmarking efforts are showing greater promise with novel SNN algorithms that align with the emerging neuromorphic architectures. This includes factors like exploiting preserved state and sparsity factors, but requires novel algorithm developments. 
While here we have considered neuromorphic computing, similar sentiments may be applied to the broader co-design of algorithm and architecture pairing. 
The counterintutive notion that great potential alone is not sufficient to stabilize co-design collaboration emphasizes the need for intentional joint action. Otherwise the less strategic approach of unilateral effort is more of a lottery where independent advances may be possible and impactful, but are subject to chance. Take for example, the recent efforts to develop architectures specialized for DNN execution. A common approach is also a common pitfall; hardware developers identify key operations of current DNN algorithms. Doing so not only offers a concrete set of operations to design to, but also speaks to an example algorithm workload of interest. However, with ensuing algorithmic developments, which the architecture may not support, this independent and incongruent development is analogous to going for the Hare in the Stag Hunt game. The known algorithm the architecture pursued is compelling, but given the time required to develop an architecture a more appealing algorithm may emerge. For example, the transformer neural network algorithm underlying modern Large Language Models (LLMs) as well as Vision Transformer (ViT) models tax architectures in ways differently than CNNs. Effectively, the best architectures for accelerating CNNs may be the pursuit of a Hare compared to the Stag like reward of being able to efficiently execute transformers and the next neural network breakthrough. 

By examining the co-design of neuromorphic computing from a game theoretic perspective, we encourage a strategic approach where algorithms and architectures are advanced in support of one another in order to provide better odds of winning the computing lottery and advancing the field of neuromorphic computing. 

\begin{ack}
Sandia National Laboratories is a multi-mission laboratory managed and operated by National Technology $\&$ Engineering Solutions of Sandia, LLC (NTESS), a wholly owned subsidiary of Honeywell International Inc., for the U.S. Department of Energy’s National Nuclear Security Administration (DOE/NNSA) under contract DE-NA0003525. This written work is authored by an employee of NTESS. The employee, not NTESS, owns the right, title and interest in and to the written work and is responsible for its contents. Any subjective views or opinions that might be expressed in the written work do not necessarily represent the views of the U.S. Government. The publisher acknowledges that the U.S. Government retains a non-exclusive, paid-up, irrevocable, world-wide license to publish or reproduce the published form of this written work or allow others to do so, for U.S. Government purposes. The DOE will provide public access to results of federally sponsored research in accordance with the DOE Public Access Plan.
\end{ack}

\bibliographystyle{unsrtnat}
\bibliography{arxiv_mlncp_neurips_nmc-game}


\end{document}